\pdfoutput=1

\documentclass[11pt]{article}

\usepackage{ACL2023}

\usepackage{times}
\usepackage{latexsym}

\usepackage{graphicx}
\usepackage{tabularx}
\usepackage{booktabs}
\usepackage{multirow}
\usepackage{setspace}
\usepackage{courier} 
\usepackage{url}            
\usepackage{booktabs}       
\usepackage{amsfonts}       
\usepackage{nicefrac}       
\usepackage{microtype}      
\usepackage{xcolor}         
\usepackage{amssymb}
\usepackage{pifont}
\usepackage[edges]{forest}

\definecolor{hidden-red}{RGB}{205, 44, 36}
\definecolor{hidden-blue}{RGB}{194,232,247}
\definecolor{hidden-orange}{RGB}{243,202,120}
\definecolor{hidden-green}{RGB}{34,139,34}
\definecolor{hidden-pink}{RGB}{255,245,247}
\definecolor{hidden-black}{RGB}{20,68,106}

\newcommand{\eg}{E.g.{,}~}%

\newcommand{\cmark}{\ding{52}}%
\newcommand{\xmark}{\ding{55}}%

\newcommand{\red}[1]{\textcolor{red}{#1}}
\newcommand{\blue}[1]{\textcolor{blue}{#1}}

\usepackage[T1]{fontenc}

\usepackage[utf8]{inputenc}

\usepackage{microtype}

\usepackage{inconsolata}

\title{A Survey of Deep Learning for Mathematical Reasoning}

\author{Pan Lu$^1$, Liang Qiu$^1$, Wenhao Yu$^2$, Sean Welleck$^{3*}$, Kai-Wei Chang$^{1*}$
\\
$^1$UCLA,
$^2$University of Notre Dame, 
$^3$University of Washington \\
\url{https://github.com/lupantech/dl4math}
}
  
\begin{document}
\maketitle



\maketitle
\renewcommand*\thefootnote{\textbf{$*$}}\footnotetext{denotes co-senior authors.}
\begin{abstract}
Mathematical reasoning is a fundamental aspect of human intelligence and is applicable in various fields, including science, engineering, finance, and everyday life. The development of artificial intelligence (AI) systems capable of solving math problems and proving theorems in language has garnered significant interest in the fields of machine learning and natural language processing. For example, mathematics serves as a testbed for aspects of reasoning that are challenging for powerful deep learning models, driving new algorithmic and modeling advances. On the other hand, recent advances in large-scale neural language models have opened up new benchmarks and opportunities to use deep learning for mathematical reasoning. In this survey paper, we review the key tasks, datasets, and methods at the intersection of mathematical reasoning and deep learning over the past decade. We also evaluate existing benchmarks and methods, and discuss future research directions in this domain.
 

\end{abstract}

\section{Introduction}
\label{sec:introduction}

``\textit{The study of mathematics, like the Nile, begins in minuteness but ends in magnificence.}''

\hfill --- Charles Caleb Colton, English writer
\break










Mathematical reasoning is a key aspect of human intelligence that enables us to comprehend and make decisions based on numerical data and language. It is applicable in various fields, including science, engineering, finance, and everyday life, and encompasses a range of abilities, from basic skills such as pattern recognition and numerical operations to more advanced skills like problem-solving, logical reasoning, and abstract thinking. The development of artificial intelligence (AI) systems capable of solving math problems and proving theorems in language has been a long-standing focus of research in the fields of machine learning and natural language processing (NLP), dating back to the 1960s \cite{feigenbaum1963computers, bobrow1964natural}. In recent years, there has been a surge of interest in this area: for instance, the number of papers has grown from approximately
10 in 2018 to 66 in 2022 (see \autoref{fig:papers} in the Appendix).

As deep learning continues to revolutionize NLP tasks such as question answering and machine translation~\cite{sutskever2014sequence, devlin2019bert}, it has also made significant strides in the field of mathematical reasoning~\cite{wang2017deep,yang2019learning,geva2020injecting,wei2022chain}. 
However, despite the impressive capabilities of these models, there is still a lack of a clear taxonomy of the different types of mathematical reasoning tasks and the specific capabilities required of deep learning models to solve them. 

Previous literature has been limited to the discussion of specific aspects, such as solving math word problems~\cite{bhattacharya2017survey,zhang2019gap,ughade2019survey}, representing numbers representation~\cite{thawani2021representing}, or solving informal problems~\cite{meadows2022survey}. 
Additionally, with the recent advancements in large language models like GPT-3~\cite{brown2020language}, there is a growing need to understand the capabilities and limitations of these models in the context of mathematical reasoning. This is where a comprehensive survey of this rapidly advancing domain becomes crucial, as it can provide an overview of the current state and limitations of the field, and indicate further research areas.


\tikzstyle{my-box}=[
    rectangle,
    draw=hidden-black,
    rounded corners,
    text opacity=1,
    minimum height=1.5em,
    minimum width=5em,
    inner sep=2pt,
    align=center,
    fill opacity=.5,
]
\tikzstyle{leaf}=[
    my-box, 
    minimum height=1.5em,
    fill=hidden-blue!90, 
    text=black,
    align=left,
    font=\normalsize,
    inner xsep=2pt,
    inner ysep=4pt,
]
\begin{figure*}[th!]
    \vspace{-2mm}
    \centering
    \resizebox{\textwidth}{!}{
        \begin{forest}
            forked edges,
            for tree={
                grow=east,
                reversed=true,
                anchor=base west,
                parent anchor=east,
                child anchor=west,
                base=left,
                font=\large,
                rectangle,
                draw=hidden-black,
                rounded corners,
                align=left,
                minimum width=4em,
                edge+={darkgray, line width=1pt},
                s sep=3pt,
                inner xsep=2pt,
                inner ysep=3pt,
                line width=0.8pt,
                ver/.style={rotate=90, child anchor=north, parent anchor=south, anchor=center},
            },
            where level=1{text width=6.2em,font=\normalsize,}{},
            where level=2{text width=10.5em,font=\normalsize,}{},
            where level=3{text width=13.5em,font=\normalsize,}{},
            where level=4{text width=12em,font=\normalsize,}{},
            [
                Deep Learning for Mathematical Reasoning, ver
                [
                    Tasks and \\ Datasets (\S \ref{sec:task})
                    [
                        Math Word Problem \\ Solving (\S \ref{sec:mwp})
                        [   
                            Textual
                            [
                                \eg MathQA~\cite{amini2019mathqa}{,}
                                SVAMP~\cite{patel2021nlp}
                                , leaf, text width=32em
                            ]
                        ]
                        [
                            Multimodal
                            [
                                \eg IconQA~\cite{lu2021iconqa}{,}
                                TabMWP~\cite{lu2022dynamic}
                                , leaf, text width=32em
                            ]
                        ]
                    ]
                    [
                        Theorem Proving (\S \ref{sec:tp})
                        [   
                            Formal
                            [
                                \eg CoqGym~\cite{yang2019learning}
                                , leaf, text width=32em
                            ]
                        ]
                        [
                            Informal
                            [
                                \eg NaturalProofs~\cite{welleck2021naturalproofs}
                                , leaf, text width=32em
                            ]
                        ]
                        [
                            Formal + Informal
                            [
                                \eg miniF2F+informal~\cite{jiang2022dsp}
                                , leaf, text width=32em
                            ]
                        ]
                    ]
                    [
                        Geometry Problem \\ Solving (\S \ref{sec:gps})
                        [
                            Without Annotations
                            [
                                \eg GEOS~\cite{seo2015solving}{,}
                                GEOS++~\cite{sachan2017textbooks}
                                , leaf, text width=32em
                            ]
                        ]
                        [
                            With Annotations
                            [
                                \eg Geometry3K~\cite{lu2021inter}{,}
                                UniGeo~\cite{chen2022unigeo}
                                , leaf, text width=32em
                            ]
                        ]
                    ]
                    [
                        Math Question \\ Answering (\S \ref{sec:mathqa})
                        [
                            Single Benchmark
                            [
                                \eg DROP~\cite{dua2019drop}{,}
                                Mathematics~\cite{saxton2019analysing}
                                , leaf, text width=32em
                            ]
                        ]
                        [
                            Unified Benchmark
                            [
                                \eg Lila~\cite{Mishra2022Lila}{,} TheoremQA~\cite{chen2023theoremqa}
                                , leaf, text width=32em
                            ]
                        ]
                    ]
                    [
                        Other Quantitative \\ Problems (\S \ref{sec:other_problem})
                        [
                            Diagram
                            [
                                \eg FigureQA~\cite{kahou2017figureqa}{,}
                                DVQA~\cite{kafle2018dvqa}
                                , leaf, text width=32em
                            ]
                        ]
                        [
                            Finance, 
                            [
                                \eg ConvFinQA~\cite{chen2022convfinqa}
                                , leaf, text width=32em
                            ]
                        ]
                        [
                            Science
                            [
                                \eg ScienceQA~\cite{lu2022learn}
                                , leaf, text width=32em
                            ]
                        ]
                        [
                            Programming
                            [
                                \eg P3~\cite{schuster2021programming}
                                , leaf, text width=32em
                            ]
                        ]
                    ]
                ]
                [
                    Deep Learning \\ Methods
                    [
                        Neural Networks (\S \ref{sec:network})
                        [
                            Seq2Seq-based  (\S \ref{sec:seq2seq})
                            [
                                \eg DNS~\cite{wang2017deep}{,}
                                AnsRat~\cite{ling2017program}
                                , leaf, text width=32em
                            ]
                        ]
                        [
                            Graph-based (\S \ref{sec:graph})
                            [
                                \eg GTS~\cite{xie2019goal}{,}
                                Graph2Tree~\cite{li2020graph}
                                , leaf, text width=32em
                            ]
                        ]
                        [
                            Attention-based  (\S \ref{sec:attention})
                            [
                                \eg Math-EN~\cite{wang2018translating}{,}
                                GROUP-ATT~\cite{li2019modeling}
                                , leaf, text width=32em
                            ]
                        ]
                        [
                            Other (\S \ref{sec:other_network})
                            [
                                \eg CNNTP~\cite{loos2017deep}{,}
                                MathDQN~\cite{wang2018mathdqn}
                                , leaf, text width=32em
                            ]
                        ]
                    ]
                    [
                        Pre-trained Language \\ Models (\S \ref{sec:pretrain})
                        [
                            Self-Supervised Learning (\S \ref{sec:self})
                            [
                                \eg GenBERT~\cite{geva2020injecting}{,}
                                Minerva~\cite{lewkowycz2022solving}
                                , leaf, text width=32em
                            ]
                        ]
                        [
                            Task-specific Fine-tuning (\S \ref{sec:finetune})
                            [
                                \eg Scratchpad~\cite{nye2021show}{,}
                                Bhaskara~\cite{Mishra2022Lila}
                                , leaf, text width=32em
                            ]
                        ]
                    ]
                    [
                        In-context Learning (\S \ref{sec:icl})
                        [
                            Example Selection (\S \ref{sec:selection})
                            [
                                \eg Few-shot-CoT~\cite{wei2022chain}{,}
                                PromptPG~\cite{lu2022dynamic}
                                , leaf, text width=32em
                            ]
                        ]
                        [
                            High-quality Chains (\S \ref{sec:chains})
                            [
                                \eg Self-Consistency~\cite{wang2022self}{,}
                                Least-to-most~\cite{zhou2022least}
                                , leaf, text width=32em
                            ]
                        ]
                    ]
                ]
            ]
        \end{forest}
    }
    \vspace{-4mm}
    \caption{Taxonomy of deep learning for mathematical reasoning. The associated tasks are elaborated in \S \ref{sec:task}, with a comprehensive dataset list found in \S \ref{sec:dataset}. Deep learning methods are further discussed in \S \ref{sec:network}, \S \ref{sec:pretrain}, and \S \ref{sec:icl}.}
    \label{fig:taxonomy}
    \vspace{-3mm}
\end{figure*}
In this paper, we survey over 180 papers from 
the NLP and AI communities in the field of deep learning for mathematical reasoning. We study various types of mathematical reasoning problems, such as math word problems, theorem proving, geometry problem solving, math question answering, and other quantitative
problems (\S\ref{sec:task}, \S\ref{sec:dataset}). Additionally, we explore different deep learning architectures for mathematical reasoning, including neural networks (\S \ref{sec:network}), pre-trained language models (\S \ref{sec:pretrain}), and recent in-context learning for large language models (\S \ref{sec:icl}).

We also analyze existing benchmarks and find that there is less focus on multi-modal and low-resource settings (\S \ref{sec:dis_datasets}). Our evidence-based studies suggest that current numeracy representations are insufficient and deep learning methods are inconsistent for mathematical reasoning (\S \ref{sec:dis_methods}). Following this, we suggest 
future research directions related to
generalization and robustness, trustworthy reasoning, learning from feedback, and multi-modal mathematical reasoning (\S \ref{sec:future}).

\section{Mathematical Reasoning Tasks}
\label{sec:task}
In this section, we briefly introduce different tasks for mathematical reasoning. A detailed summary and discussion of commonly used datasets can be found in \autoref{tab:datasets} and Appendix \ref{sec:dataset}.

\noindent\textbf{Math Word Problem Solving}.  Developing algorithms to automatically solve math word problems (MWPs) has been of interest to NLP researchers for decades \cite{feigenbaum1963computers,bobrow1964natural}. An example of a MWP is shown in \autoref{tab:mwp}. A question involves four basic arithmetic operations with single or multiple operation steps. The challenge posed by MWPs lies in the need for language comprehension, semantic parsing, and the application of multiple mathematical reasoning skills.


\renewcommand{\arraystretch}{1.4}
\begin{table}[t!]
\small
\centering
\begin{tabular}{ |p{6.5cm}| } 
 \hline
 \textbf{Question:} Bod has 2 apples and David has 5
apples. How many apples do they have in total?\\ 
 \hline
 \textbf{Rationale:} $x = 2 + 5$ \\ 
 \hline
 \textbf{Solution:} $7$ \\ 
 \hline
\end{tabular}
\vspace{-2mm}
\caption{A typical math word problem.}
\vspace{-3mm}
\label{tab:mwp}
\end{table}
\renewcommand{\arraystretch}{1.0}

\noindent\textbf{Theorem Proving}. Automating theorem proving is a long-standing challenge in AI~\cite{newell1957empirical,feigenbaum1963computers}.
The problem is to demonstrate the truth of a mathematical claim (a \textit{theorem}) through a sequence of logical arguments (a \textit{proof}).
Theorem proving tests various skills, such as choosing effective multi-step strategies, using background knowledge, and performing symbolic manipulations.

\begin{figure}[t!] 
\centering
\includegraphics[width=0.49\textwidth]{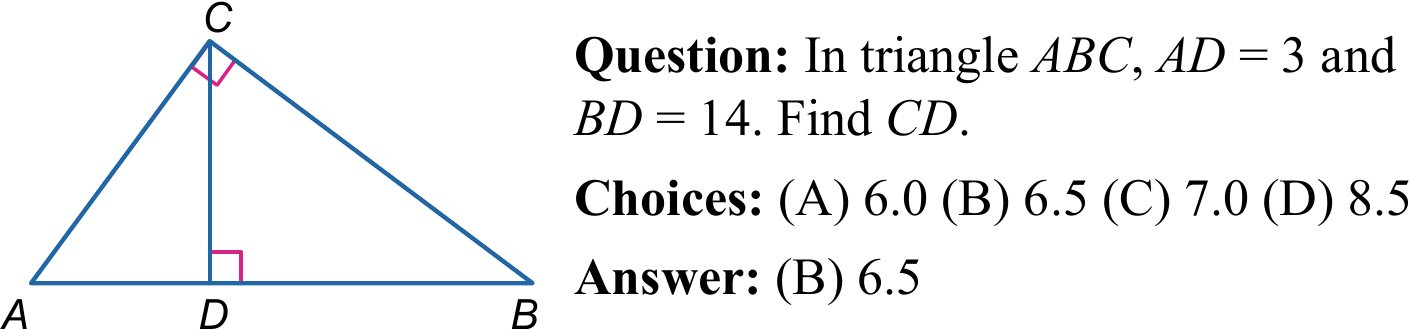}
\vspace{-4mm}
\caption{An example of geometry problems.}
\vspace{-3mm}
\label{fig:gps}
\end{figure}

\noindent\textbf{Geometry Problem Solving}. Automated geometry problem solving (GPS) is also a long-standing mathematical reasoning task \cite{gelernter1960empirical,wen1986basic}. As shown in \autoref{fig:gps}, a geometry problem consists of a textual description and a diagram. The multimodal inputs describe the entities, attributes, and relationships of geometric elements, and the goal is to find the numeric solution to an unknown variable. 

\noindent\textbf{Math Question Answering}. There is a wide range of question answering (QA) benchmarks that center around mathematical reasoning, which we refer to as math question answering (MathQA). For example, DROP~\cite{dua2019drop} is a MathQA dataset that requires discrete reasoning
to answer questions such as ``Which kicker kicked the most field goals?'' over the content of paragraphs.


\section{Neural Networks for Mathematical Reasoning}
\label{sec:network}
Neural networks have become a popular tool in the field of mathematical reasoning, mirroring their success in NLP. In recent years, a number of different neural network architectures have been proposed for mathematical reasoning tasks, including Seq2Seq-based networks, graph-based networks, and attention-based networks. These methods are outlined in more detail in \autoref{tab:network} in the Appendix.

\subsection{Seq2Seq-based Networks for Math}
\label{sec:seq2seq}

Sequence-to-sequence (Seq2Seq) \cite{sutskever2014sequence} neural networks have been successfully applied to mathematical reasoning tasks, such as math word problem solving \cite{wang2017deep}, theorem proving ~\cite{yang2019learning}, geometry problem solving~\cite{robaidek2018data}, and math question answering \cite{tafjord2019quarel}. 
A Seq2Seq model uses an encoder-decoder architecture and usually formalizes mathematical reasoning as a sequence generation task. The basic idea behind this approach is to map an input sequence (e.g. a mathematical problem) to an output sequence (e.g. an equation, program, and proof). Common encoders and decoders include Long Short Term Memory network (LSTM)~\cite{hochreiter1997long}, Gated Recurrent Unit (GRU)~\cite{cho2014learning}, and their bidirectional variants: BiLSTM and BiGRU. 
A large amount of work has shown the performance advantage of Seq2Seq models over previous statistical learning approaches~\cite{ling2017program, wang2018translating, huang2018neural, wang2019template, li2019modeling}.

\subsection{Graph-based Networks for Math}
\label{sec:graph}
Seq2Seq approaches show their advantages of generating mathematical expressions without relying on hand-crafted features. It is noteworthy that mathematical expressions can be represented as tree-based structures, such as abstract syntax trees (ASTs) and graph-based structures, which capture the structural information in the expressions. However, Seq2Seq methods do not explicitly this important information. To address this limitation, graph-based neural networks have been developed to explicitly model the structure within expressions.
Sequence-to-tree (Seq2Tree) models explicitly model the tree structure when encoding the output sequences~\cite{xie2019goal,wu2020knowledge,zaporojets2021solving,qin2021neural}.
For example, ~\citet{liu2019tree} devise a Seq2Tree model to better use information from an equation's AST. Seq2DAG \cite{cao2021bottom}, instead, applies a sequence-to-graph (Seq2Graph) framework when generating the equations since the graph decoder is able to extract complex relationships among multiple variables. The graph-based information can also be embedded when encoding the input mathematical sequences \cite{zhang2020graph,shen2020solving,li2020graph,wu2021edge}. 

\subsection{Attention-based Networks for Math}
\label{sec:attention}
The attention mechanism has been successfully applied to NLP \cite{bahdanau2015neural} and vision problems \cite{xu2015show,woo2018cbam}, taking into account the hidden vectors of the inputs during the decoding processing. Recently, researchers have been exploring its usefulness in mathematical reasoning tasks, as it can be used to identify the most important relationships between mathematical concepts. For instance, MATH-EN~\cite{wang2018translating} is a math word problem solver which benefits from long-distance dependency information learned by self-attention. Attention-based methods have also been applied to other mathematical reasoning tasks such as geometry problems solving \cite{robaidek2018data,chen2021geoqa} and theorem proving \cite{yang2019learning}. Various attention mechanisms have been studied to extract better representations, such as Group-ATT \cite{li2019modeling} which uses different multi-head attention to extract various types of MWP features, and graph attention which is applied to extract knowledge-aware information in \cite{wu2020knowledge}.

\subsection{Other Neural Networks for Math}
\label{sec:other_network}

Deep learning approaches to mathematical reasoning tasks can also make use of other neural networks, such as convolutional neural networks (CNN) and multimodal networks. Some work encodes the input text using a convolutional neural network architecture, giving the model the ability to capture long-term relationships between symbols in the input \cite{gehring2017convolutional, wang2018translating, wang2018translating, robaidek2018data, alemi2016deepmath, loos2017deep}. For example, the first application of deep neural networks for  theorem proving is proposed in \cite{alemi2016deepmath}, which relies on convolutional networks for premise selection.

Multimodal mathematical reasoning tasks, such as geometry problem solving and diagram-based mathematical reasoning, are formalized as visual question answer (VQA) problems \cite{kafle2018dvqa,chen2021geoqa,lu2021iconqa}. In this domain, visual inputs are encoded using ResNet \cite{he2016deep} or Faster-RCNN \cite{ren2015faster}, while textual representations are obtained via GRU or LTSM. Subsequently, the joint representation is learned using multimodal fusion models, such as BAN \cite{Kim2018}, FiLM \cite{perez2018film}, and DAFA \cite{gao2019dynamic}.

Other deep neural network structures can also be used in mathematical reasoning. A Graph Neural Network (GNN) is employed for geometry problem parsing in \citet{zhang2022learning}, taking advantage of its success in spatial reasoning. WaveNet has been applied to theorem proving \cite{loos2017deep, bansal2019holist}, due to its ability to address longitudinal time-series data. Furthermore, Transformers are found to outperform GRU in generating mathematical equations in DDT \cite{meng2019solving}. Finally, MathDQN \cite{wang2018mathdqn} is the first work to explore reinforcement learning for math word problem solving, taking advantage of its strong search capabilities.

\section{Pre-trained Language Models for Mathematical Reasoning}
\label{sec:pretrain}

\begin{table*}[th!]
\centering
\fontsize{8.0pt}{\baselineskip}\selectfont 
\renewcommand\tabcolsep{7pt} 
\renewcommand\arraystretch{0.7} 
\begin{tabular}{lccccccc} 
\toprule
\textbf{Paper} & \textbf{Backbone} & \textbf{Size} & \textbf{Corpus} & \textbf{Pre-training task} \\ 
\midrule
\textbf{GPT-\textit{f}}~\cite{polu2020generative} & Transformer~\citeyearpar{vaswani2017attention} & 774M & Math & Causal language modeling \\ 
\textbf{LISA}~\cite{jiang2021lisa} & Transformer~\citeyearpar{vaswani2017attention} & 163M & Math & Causal language modeling \\
\textbf{MATH-PLM}~\cite{hendrycks2021measuring} & GPT-2~\citeyearpar{radford2019language} & 1.5B & Math & Causal language modeling \\
\textbf{MWP-BERT}~\cite{liang2022mwp} & RoBERTa~\citeyearpar{liu2019roberta} & 123M & Math & 8 numeracy augmented tasks \\
\textbf{TaPEx}~\cite{liu2022tapex} & BART~\citeyearpar{lewis-etal-2020-bart} & 406M & SQL & Query result generation \\
\textbf{HTPS}~\cite{lample2022hypertree} & Transformer~\citeyearpar{vaswani2017attention} & 600M & Math & Masked Seq2Seq modeling \\
\textbf{Thor}~\cite{jiang2022thor} & Transformer~\citeyearpar{vaswani2017attention} & 700M & Github, arXiv & Causal language modeling \\
\textbf{PACT}~\cite{han2021proof} & Transformer~\citeyearpar{vaswani2017attention} & 837M & Math & Masked/Causal language modeling \\
\textbf{Minerva}~\cite{lewkowycz2022solving} & PaLM~\citeyearpar{chowdhery2022palm} & 540B & Science \& Math & Causal language modeling \\
\midrule
\textbf{GenBERT}~\cite{geva2020injecting} & BERT~\citeyearpar{devlin2019bert} & 110M & Number, Text & Masked/Causal language modeling \\
\textbf{NF-NSM}~\cite{feng2021injecting} & RoBERTa~\citeyearpar{liu2019roberta} & 110M & Number & Number prediction \\
\textbf{LIME}~\cite{wu2021lime} & Transformer~\citeyearpar{vaswani2017attention} & 11B & Math & Causal language modeling \\
\textbf{Set}~\cite{wu2022insights} & T5~\citeyearpar{raffel2020exploring} & 60M & Math & Unique token generation \\
\bottomrule
\end{tabular}
\vspace{-2mm}
\caption{Comparison of pre-training language models for mathematical reasoning.}
\vspace{-2mm}
\label{tab:pre-trained_lm}
\end{table*}

Pre-trained language models~\cite{devlin2019bert,radford2019language,brown2020language} have demonstrated remarkable performance gains on a wide range of NLP tasks.
By pre-training on a large corpus of text, the models learn valuable world knowledge~\cite{guu2020retrieval}, which could be applied to downstream tasks.
Similar ideas can be applied to math-related problems, and previous work has shown the promising performance of pre-trained language models in
answering math word problems~\cite{kim2020point}, assisting with theorem proving~\cite{wu2022autoformalization}, as well as solving other mathematical tasks~\cite{charton2021linear}.

However, though large language models excel in modeling natural language, there are several challenges to using them for mathematical reasoning. First, pre-trained language models are not specifically trained on mathematical data. This likely contributes to them being less proficient in math-related tasks compared to natural language tasks. There is also less mathematical or scientific data available for large-scale pre-training compared to text data.
Second, the size of pre-trained models continues to grow, making it expensive to train the entire model from scratch for specific downstream tasks.
Additionally, downstream tasks may deal with different input formats or modalities, such as structured tables~\cite{zhao2022multihiertt} or diagrams~\cite{lu2021iconqa}. To address these challenges, researchers have to adjust pre-trained models by finetuning them on downstream tasks or adapting the neural architectures.


\subsection{Self-Supervised Learning for Math}
\label{sec:self}
Self-supervised learning is a machine learning approach in which an algorithm learns to perform a task without being explicitly provided with labeled training data. 
\autoref{tab:pre-trained_lm} provides a list of language models pre-trained with self-supervised tasks for mathematical reasoning.

\noindent\textbf{Model scale.}
There is a clear trend that pre-trained language models have become increasingly larger in the past few years~\cite{devlin2019bert, lewis-etal-2020-bart, raffel2020exploring, radford2019language, brown2020language}. A recent study~\cite{liang2022holistic} shows that model scale within a model family reliably predicts model accuracy. The study also mentions an interesting thresholding effect: ``all models that win head-to-head model comparisons for accuracy at a rate well above chance are at least 50B parameters''. 
A similar size-growing trend can be observed in the field of mathematical reasoning with pre-trained language models. For example, MWP-BERT~\cite{liang2022mwp} uses a backbone of BERT (110M)~\cite{devlin2019bert} and RoBERTa (123M)~\cite{liu2019roberta} for Math Word Problems. 
Most recently, Minerva~\cite{lewkowycz2022solving}, which is based on the PaLM~\cite{chowdhery2022palm} pre-trained language model, has a size up to 540B parameters.

\noindent\textbf{Pre-training corpus.}
There are generally two types of pre-training corpus for mathematical language models. 
(i) Curated datasets from openly accessible sources. 
For example, \citet{hendrycks2021measuring} present the first large-scale mathematics pre-training dataset with step-by-step solutions in natural language and \LaTeX, called the Auxiliary Mathematics Problems and Solutions (AMPS). AMPS consists of Khan Academy and Mathematica data. Minerva~\cite{lewkowycz2022solving} collects a high-quality dataset containing scientific and mathematical data, which contains 38.5B tokens from webpages filtered for mathematical content and from papers submitted to the arXiv preprint server. Thor~\cite{jiang2022thor} pre-trains a language model on the GitHub + arXiv subsets of The Pile~\cite{gao2020pile}.
(ii) Synthetic datasets based on templates or interaction with engines. Recent work~\cite{wu2021lime,krishna2021does,ri2022pretraining,anderson2022improving,wu2022insights} shows that pre-training on data that is fully synthetically generated—synthetic pre-training can actually provide substantial gains. Representative work includes TaPEX~\cite{liu2022tapex}, which obtains a pre-training corpus by automatically synthesizing executable SQL queries and their execution outputs. LISA~\cite{jiang2021lisa} extracts lemmas and theorems by interacting with the Isabelle standard library and the Archive of Formal Proofs. GenBERT~\cite{geva2020injecting} generates numerical and textual pre-training datasets based on manually crafted and extracted templates. 

\noindent\textbf{Pre-training tasks.}
General pre-training language models have two typical self-supervised learning tasks: (i) Masked Language Modeling (MLM), where it randomly masks a portion of words in each sequence to predict the outcome; (ii) Causal Language Modeling (CLM), where the model is trained to predict the next token in a sequence of tokens. Following the same paradigm, researchers pre-train language models with MLM and CLM tasks on mathematical or scientific corpora for downstream tasks~\cite{polu2020generative,hendrycks2021measuring,han2021proof,jiang2022thor}. 

There is also recent work that designs customized tasks to inject mathematical reasoning capabilities into language models. For instance, \citet{liang2022mwp} pre-train language models with a suite of 8 numeracy-augmented tasks with consideration of reasoning logic and numerical properties. LIME~\cite{wu2021lime} proposes synthetic pre-training tasks to learn three reasoning primitives: deduction, induction, and abduction before learning more complex reasoning skills, which also be regarded as a form of curriculum learning. 
\begin{table}[t!]
\centering
\fontsize{8.0pt}{\baselineskip}\selectfont 
\renewcommand\tabcolsep{6pt} 
\renewcommand\arraystretch{0.7} 
\begin{tabular}{lccccccc} 
\toprule 
\textbf{Paper} & \textbf{Backbone} & \textbf{Task} \\ 
\midrule
\textbf{EPT}~\citeyearpar{kim2020point} & ALBERT~\citeyearpar{lan2019albert} & MWP \\
\textbf{Generate \& Rank}~\citeyearpar{shen2021generate} & BART~\citeyearpar{lewis-etal-2020-bart} & MWP \\
\textbf{RPKHS}~\citeyearpar{yu2021improving} & RoBERTa~\citeyearpar{liu2019roberta} & MWP\\
\textbf{PatchTRM}~\citeyearpar{lu2021iconqa} & ResNet+BERT~\citeyearpar{devlin2019bert} & MWP \\ 
\textbf{GSM8K-PLM}~\citeyearpar{cobbe2021training} & GPT-3~\citeyearpar{brown2020language} & MWP \\
\textbf{BERT-TD+CL}~\citeyearpar{li2022seeking} & BERT~\citeyearpar{devlin2019bert} & MWP \\
\textbf{DeductReasoner}~\citeyearpar{jie2022learning} & RoBERTa~\citeyearpar{liu2019roberta} & MWP \\
\textbf{Self-Sampling}~\citeyearpar{ni2022learning} & GPT-Neo~\citeyearpar{gao2020pile} & MWP \\
\textbf{Bhaskara}~\citeyearpar{Mishra2022Lila} & GPT-Neo~\citeyearpar{gao2020pile} & MWP \\
\midrule
\textbf{miniF2F-PLM}~\citeyearpar{zheng2021minif2f} & GPT-\textit{f}~\citeyearpar{polu2020generative} & TP \\
\textbf{NaturalProver}~\citeyearpar{welleck2022naturalprover} & GPT-3~\citeyearpar{brown2020language} & TP \\
\midrule
\textbf{Inter-GPS}~\citeyearpar{lu2021inter} & BART~\citeyearpar{lewis-etal-2020-bart} & GPS \\
\textbf{UniGeo}~\citeyearpar{chen2022unigeo} & VL-T5~\citeyearpar{pmlr-v139-cho21a} & GPS \\
\textbf{DPE-NGS}~\citeyearpar{cao2022augmented} & RoBERTa~\citeyearpar{liu2019roberta} & GPS \\
\midrule
\textbf{Aristo}~\citeyearpar{clark2020f} & RoBERTa~\citeyearpar{liu2019roberta} & MathQA \\
\textbf{FinQANet}~\citeyearpar{chen2021finqa} & RoBERTa~\citeyearpar{liu2019roberta} & MathQA\\
\textbf{TAGOP}~\citeyearpar{zhu2021tat} & RoBERTa~\citeyearpar{liu2019roberta} & MathQA \\
\textbf{MT2Net}~\citeyearpar{zhao2022multihiertt} & RoBERTa~\citeyearpar{liu2019roberta} & MathQA\\
\midrule
\textbf{Scratchpad}~\citeyearpar{nye2021show} & Transformer~\citeyearpar{vaswani2017attention} & Mixed\\
\textbf{LAMT}~\citeyearpar{charton2021linear} & Transformer~\citeyearpar{vaswani2017attention} & Mixed \\ 
\bottomrule
\end{tabular}
\vspace{-2mm}
\caption{Finetuned pre-trained language models for downstream mathematical reasoning tasks.}
\vspace{-3mm}
\label{tab:finetune}
\end{table}

\subsection{Task-specific Fine-tuning for Math}
\label{sec:finetune}
Task-specific fine-tuning is a technique to improve the performance of a pre-trained language model on a specific task. This is also a common practice when there is not enough data for training the large models from scratch.  As shown in \autoref{tab:finetune}, existing work fine-tunes pre-trained language models on a variety of downstream tasks, such as math word problems~\cite{kim2020point,shen2021generate}, MathQA~\cite{zhao2022multihiertt}, geometry problem solving~\cite{lu2021inter}, linear algebra~\cite{charton2021linear}, and theorem proving~\cite{welleck2022naturalprover}. Apart from fine-tuning the model parameters, some work also uses pre-trained language models as encoders and ensembles them with other modules for downstream tasks~\cite{lu2021iconqa}.

\section{In-context Learning for Mathematical Reasoning}
\label{sec:icl}

\begin{table*}[th!]
\vspace{-3mm}
\centering
\fontsize{8.5pt}{\baselineskip}\selectfont 
\renewcommand\tabcolsep{5.0pt} 
\renewcommand\arraystretch{0.75} 
\begin{tabular}{lcccccc} 
\toprule
\multirow{2}{*}{\textbf{Models}} & \textbf{Engine} & \textbf{ICL} & \textbf{Rationale} & \textbf{Rationale} & \multirow{2}{*}{\textbf{Post method}}
 \\ 
& \textbf{(best performed)} & \textbf{source} & \textbf{type} & \textbf{source} &
 \\ 
\midrule
Few-shot-CoT~\cite{wei2022chain}  & PaLM (540B) & Random & Language & Hand-crafted & - \\
Self-Consistency-CoT~\cite{wang2022self} & Codex  (175B) & Random & Language & Hand-crafted & Self-consistency \\
Least-to-most CoT~\cite{zhou2022least}  & Codex (175B) & Random & Language & Hand-crafted & - \\
PromptPG-CoT~\cite{lu2022dynamic}  & GPT-3 (175B) & RL & Language & Hand-crafted & - \\
Retrieval-CoT~\cite{zhang2022automatic}  & GPT-3 (175B) & Retrival & Language & Auto-generated & - \\
Auto-CoT~\cite{zhang2022automatic} & Codex (175B) & Clustering & Language & Auto-generated & - \\
Complexity-CoT~\cite{fu2022complexity} & GPT-3 (175B) & Complexity & Language & Hand-crafted & Self-consistency \\
Few-shot-PoT~\cite{chen2022program} & GPT-3 (175B) & Random & Code & Hand-crafted & - \\ 
\bottomrule
\end{tabular}
\vspace{-2mm}
\caption{In-context learning with large language models for mathematical reasoning. For GPT-3, all papers use the $\mathrm{text}$-$\mathrm{davinci}$-$\mathrm{002}$ version; for Codex, all papers use the $\mathrm{code}$-$\mathrm{davinci}$-$\mathrm{002}$. RL is short for reinforcement learning.
}
\vspace{-3mm}
\label{tab:incontext}
\end{table*}

Large language models (LLMs), such as GPT-3~\citep{brown2020language}, have recently revolutionized the field of natural language processing (NLP), especially on account of their powerful few-shot in-context learning capabilities~\cite{brown2020language}. 
In-context Learning (ICL) enables LLMs to perform target tasks by providing some task examples as conditions at inference time, without updating model parameters~\cite{radford2019language,brown2020language}. 
ICL allows users to quickly build models for new use cases without worrying about fine-tuning and storing a large amount of new parameters for each task, so it is widely used in few-shot settings nowadays~\cite{min2022rethinking}.

An in-context example typically contains an input-output pair with some prompt words, e.g., \textit{Please select the largest number from the list. Input: [2, 4, 1, 5, 8]. Output: 8}, and few-shot works by giving multiple examples, and then a final input example, where the model is expected to predict the output. 
However, such standard few-shot promptings, in which the LLM is given in-context examples of input–output pairs in front of test-time examples, have not yet proved sufficient to achieve high performance on challenging tasks such as mathematical reasoning~\cite{rae2021scaling}.

Chain-of-thought prompting (CoT) \cite{wei2022chain} leverages intermediate natural language rationales as prompts to enable LLMs to first generate \textit{reasoning chains} and then predict an answer for an input question. For example, a CoT prompt for solving the math word problem could be

\begin{quote}
 \textbf{Question:} Roger has 5 tennis balls. He buys
2 more cans of tennis balls. Each can
has 3 tennis balls. Then, how many tennis
balls does Roger have now? \\
\textbf{Answer:} \textit{Roger started with 5 balls. 2 cans
of 3 tennis balls each are 6 tennis
balls. 5 + 6 = 11.} The answer is \underline{11}.
\end{quote}

Apart from \citet{kojima2022large} showing that LLMs are decent zero-shot reasoners when given the ``Let's think step by step!'' prompt, most of the recent work has focused on how to improve chain-of-thought reasoning under the few-shot setting.
This work is mainly divided into two parts, (i) selecting better in-context examples and (ii) creating better reasoning chains.

\subsection{In-context Example Selection}
\label{sec:selection}

Early chain-of-thought work randomly or heuristically selects in-context examples. However, recent studies have shown that this type of few-shot learning can be highly unstable across different selections of in-context examples~\cite{rubin2021learning,liu2022makes}.
Therefore, which in-context reasoning examples make the most effective prompts is still an unknown problem in the literature. To address the limitation, recent work has investigated various methods to optimize the in-context examples selection process~\cite{rubin2021learning,zhang2022automatic,lu2022dynamic,yu2022generate,fu2022complexity}. For example, 
\citet{rubin2021learning} attempt to address this issue by retrieving semantically similar examples. 
In addition, \citet{fu2022complexity} propose complexity-based prompting, which chooses examples with complex reasoning chains, i.e., chains with more reasoning steps, as the prompt. PromptPG~\cite{lu2022dynamic} learns to select optimal in-context examples via reinforcement learning (RL) from a candidate pool.

\subsection{High-quality Reasoning Chains}
\label{sec:chains}
Early chain of thought work (e.g., \citet{wei2022chain}) mainly relies on a single human-annotated reasoning chain as a prompt. 
However, manually creating reasoning chains has two disadvantages. 
First, as tasks become more complex, current models may not be sufficient to learn to perform all necessary reasoning steps and cannot easily generalize to different tasks.
Second, a single decoding process is vulnerable to incorrect inference steps, leading to an incorrect prediction as the final answer.
To address this limitation, recent studies mainly focus on two aspects, (i) hand-crafting more complex demonstrations, which we refer to as \textit{process-based approaches}~\cite{zhou2022least,chen2022program}, (ii) leveraging ensemble-like methods, which we refer to as \textit{outcome-based approaches}~\cite{wang2022self,li2022advance}. 

\vspace{0.05in}
\noindent\textbf{Process-based approaches} aim to improve the chain-of-thought reasoning quality, especially for complex reasoning tasks.
In least-to-most prompting~\cite{zhou2022least}, the problem-solving process is implemented through two-stage prompting: (i) reducing a complex problem into a list of sub-problems; (ii) solving these sub-problems sequentially, so that solving a given sub-problem is facilitated by the answers to previously solved sub-problems. Similarly, \citet{khot2022decomposed} leverage diverse decomposition structures and use different prompts to answer each sub-question. Apart from these multi-step reasoning methods, \citet{chen2022program,gao2022pal} propose program-of-thoughts (PoT), an alternative solution that uses large language models to express the reasoning process as a program. The computation is then relegated to an external computer, which executes the generated programs to derive the answer. A more recent work, Chameleon \cite{lu2023chameleon}, integrates different tools to enhance the abilities of LLMs for compositional reasoning.

\vspace{0.05in}
\noindent\textbf{Outcome-based approaches} acknowledge the potential incorrectness of an individual reasoning path, and instead use multiple reasoning paths~\cite{wang2022self,li2022advance}.
Self-consistency~\cite{wang2022self} generates a set of reasoning paths by sampling from the language model, and marginalizes out the reasoning paths  by choosing the most common answer.
In addition to using sampling with a single prompt to produce multiple reasoning paths, \citet{li2022advance} propose to introduce diverse prompts through ``self-teaching'', as a complementary solution to produce a higher degree of diversity.

\section{Discussion and Findings}
\label{sec:discussion}

\subsection{Analysis of Benchmarks}
\label{sec:dis_datasets}

\textbf{The multi-modal setting is underexplored but is gaining increasing attention.} Most existing benchmarks for mathematical reasoning have targeted the textual-only modality. However, visual elements can provide a rich source of quantitative information, making multi-modal datasets beneficial for reasoning over quantitative relations in natural images \cite{lu2022learn}, abstract diagrams \cite{lu2021iconqa}, figures \cite{kahou2017figureqa}, and charts \cite{kafle2018dvqa}. Tables, which are commonly found in daily documents and contain hierarchically structured information, have also been the focus of tasks that require quantitative reasoning over textual and tabular context \cite{chen2021finqa,zhu2021tat,zhao2022multihiertt,lu2022dynamic}. In addition, recent datasets have been developed for mathematical reasoning grounded on conversations \cite{sun2019dream,zhang2021noahqa,chen2022convfinqa}, as well as reports \cite{chen2022convfinqa}.

\vspace{0.05in}
\noindent \textbf{Pioneering work is emerging in the exploration of low-resource settings.} Despite the creation of various datasets, mathematical reasoning in low-resource settings remains largely under-explored. Pioneering research has developed mathematical reasoning benchmarks for financial \cite{chen2021finqa,zhu2021tat,zhao2022multihiertt} and scientific domains \cite{lu2022learn}. Additionally, there have been attempts to build non-English datasets for Chinese \cite{wang2017deep,qin2020semantically,yu2021geore} and Arabic \cite{alghamdi2022armath} for mathematical reasoning.

\vspace{0.05in}
\noindent \textbf{Diverse rationale annotations have been widely explored.} Complex reasoning usually involves multiple steps to arrive at the final answer. To bridge this gap, datasets annotated with intermediate rationales such as logic forms \cite{tafjord2019quarel,lu2021inter}, programs \cite{amini2019mathqa,chen2021finqa,chen2021geoqa,cao2022augmented,chen2022unigeo}, and reasoning graphs \cite{zhang2021noahqa} have been proposed to train models for complex reasoning tasks. Python programs are used as reasoning annotations in \cite{austin2021program, Mishra2022Lila} due to their enhanced accessibility and readability. To imitate the reasoning process of a human, a more recent trend is to annotate solutions in natural language \cite{ling2017program,cobbe2021training,lu2022dynamic,hendrycks2021measuring,lu2022learn}.

\subsection{Analysis of Deep Learning Methods}
\label{sec:dis_methods}


\begin{table}[t!]
\vspace{-3mm}
\centering
\fontsize{7.5pt}{\baselineskip}\selectfont 
\renewcommand\tabcolsep{1.0pt} 
\renewcommand\arraystretch{0.70} 
\begin{tabular}{lcccc} 
\toprule 
\textbf{} & \textbf{T5} & \textbf{UnifiedQA} & \textbf{GPT-3} & \textbf{GPT-3} \\ 
 & (Large) & (Large) & (davinci-002) & (davinci-003) \\ 
\midrule
3 balls + 5 balls = & \xmark & \red{5 balls} & 8 balls & 8 balls \\
23 balls + 145 balls = & \xmark & \xmark & \red{58 balls} & 168 balls \\
23 balls + 1,855 balls = & \xmark & \xmark & \red{2,878 balls}  & \red{2,988 balls} \\
\bottomrule
\end{tabular}
\vspace{-2mm}
\caption{Language models struggle with large numbers.}
\vspace{-4mm}
\label{tab:number}
\end{table}

\textbf{Is the current representation of numeracy sufficient?} 
The standard practice for deep learning techniques is to treat numbers in the same way as words. Early neural network methods create a vocabulary that maps input words and numbers to token IDs, resulting in less frequent numbers being collapsed into an ``\texttt{UNK}'' token. Recent language models use subword tokenization techniques \cite{wu2016google,sennrich2016neural} to split numbers into atomic tokens.
Recent studies have shown that these tokenization approaches are suboptimal \cite{wallace2019nlp,lin2020birds,zhang2020language,zhang2022estimating}.

Two numbers on the same or close number line could have surface forms with no shared common tokens. For example, a number like $1598$ is tokenized as ``$15$'' and ``$98$'' in GPT-3, while another format like $1,598$ is split as three different tokens: ``$1$'', ``$,$'', and ``$598$''. This lack of consistent representation can make it difficult for deep learning models to effectively process numbers, especially when compared to pure text. The insufficient  representations of numbers can lead to out-of-distribution (OOD) problems. \autoref{tab:number} provides examples of where language models tend to struggle with large numbers.  Although increasing model scales could help, even the state-of-the-art large language model GPT-3 performs poorly when reasoning over large numbers. Some recent work suggests that using scientific notation \cite{zhang2020language} and digit-level decomposition \cite{geva2020injecting} may be helpful in improving numeracy representation, but this remains an open problem.

\begin{table}[t!]
\vspace{-3mm}
\begin{spacing}{0.8}
\centering
\fontsize{8.0pt}{\baselineskip}\selectfont 
\renewcommand\tabcolsep{3.5pt} 
\renewcommand\arraystretch{0.70} 
\begin{tabular}{p{0.27\textwidth}p{0.18\textwidth}} 
\toprule 
\textbf{Problems} & \textbf{GPT-3} (text-davinci-002) \\
\midrule
John had 8 balls and he gave 3 to Mary. How many balls does John have now? & John has 5 balls.  \\
\midrule
\blue{John had 3 apples.} John had 8 balls and he gave 3 to Mary. How many balls does Mary have now? & \red{Mary has 5 balls.} \\
\midrule
John had 8 balls and he gave 3 to Mary. Who has more balls now? & John has more balls.  \\
\midrule
John had 8 balls and he gave 3 to Mary. \blue{Does John have more balls now?} & \red{No, John has 5 balls now.}  \\
\midrule
John had 8 balls and he gave 4 to Mary. Does John have more balls now? & No, John has 4 balls now.  \\
\midrule
John had 8 balls and he gave 4 to Mary. \blue{Who has more balls now?} & \red{John has more balls.}  \\
\bottomrule
\end{tabular}
\vspace{-2mm}
\caption{Examples where large language models are not consistent for mathematical reasoning.}
\vspace{-3mm}
\label{tab:consistency}
\end{spacing}
\end{table}

\vspace{0.05in}
\noindent \textbf{Are deep learning methods consistent for mathematical reasoning?} Recent developments in deep learning have led to impressive results on various mathematical reasoning tasks. The zero-shot-CoT Minerva 540B achieves a score of 75.0\% on the MMLU-STEM benchmark \cite{hendrycks2020measuring}, which assesses multitask reasoning ability in the fields of science, technology, engineering, and mathematics (STEM) at both high school and college levels. Similarly, few-shot-CoT GPT-3 175B achieves a high accuracy of 93.0\% on the MultiArith task. However, the question remains as to whether these methods are sufficiently advanced to tackle more complex problems.

There is strong evidence that deep learning methods for mathematical reasoning are not robust and susceptible to adversarial attacks \cite{lin2020birds,patel2021nlp,mishra2022numglue,Mishra2022Lila,welleck2022symbolic}. The SVAMP \cite{patel2021nlp} dataset is a collection of one-unknown arithmetic word problems up to grade 4, with slight word variations from previous datasets.
It is surprising that current state-of-the-art (SOTA) methods perform poorly on this dataset, with Graph2Tree achieving only a 43.8\% accuracy and zero-shot-CoT GPT-3 (175B)  only reaching  63.7\%, which is just above an ``F'' grade. \autoref{tab:consistency} also shows the inconsistent performance of the zero-shot GPT-3 model in scenarios with slightly different descriptions, while human performance remains unchanged. This indicates a lack of consistency in the mathematical reasoning ability of SOTA large language models.

\section{Future Work}
\label{sec:future}






\subsection{Generalization and Robustness}
Despite impressive progress, neural models commonly display generalization and robustness failures on reasoning tasks. 
For example, above we discussed difficulties in generalizing to larger numbers (\autoref{tab:number}) or remaining robust to nearby problems (\autoref{tab:consistency}), while others identify failures in generalizing to longer problems than those observed in training (e.g., \citet{anil2022exploring}).
One direction is to explore new inference-time~\cite{Jung2022MaieuticPL,mitchell2022enhancing} or fine-tuning~\cite{anil2022exploring} strategies.

Another aspect of generalization relates to the role of \textit{memorization}.
For example, is the ability to produce a complex solution dependent on seeing many similar solutions during training, or even on memorizing the solution? 
Term frequency in the pretraining corpus is known to impact accuracy in simple arithmetic tasks~\cite{razeghi2022ImpactOP} or factual question answering~\cite{kandpal2022LargeLM}. 
On the other hand, \citet{lewkowycz2022solving} did not find evidence of memorization in complex outputs, yet their training set and model are not available for inspection.
Gaining a full understanding of these factors for complex problems and outputs (e.g., multi-step solutions or proofs) requires more analysis, as well as accessible datasets and models.




\subsection{Trustworthy Reasoning}


Recent advances in language models have demonstrated their powerful capabilities for mathematical reasoning. However, due to the potential for generating ungrounded answers~\cite{nakano2021webgpt}, users can't always trust the predicted outcomes or have to verify then with extra efforts. Even with recent prompting strategies that provide rationales before making predictions~\cite{wei2022chain}, language models can still hallucinate statements, produce flawed reasoning,  and output wrong answers.
Consequently, novel approaches that enable more reliable reasoning are needed urgently.
Some potential directions for this include: (i) using language models to provide evidence, such as theorems, to support the reasoning process; (ii) incorporating a mechanism that makes a judgment when the model is unsure of the answer; and (iii) using a model itself or another module to detect and locate mistakes in a model's reasoning.

\subsection{Learning from Feedback}

Another important direction to further improve language models for mathematical reasoning is to let the model learn from feedback. Such a process makes the continual improvement of models' output quality and safety possible. 
An example is using 
reinforcement learning from human feedback (RLHF)~\cite{ouyang2022training} to align language models with instructions. The idea is to let humans rank the generated outputs of language models and use the learned reward function to finetune the language model with policy gradient~\cite{ouyang2022training,glaese2022improving,qiu2022valuenet}. 
In the context of mathematical reasoning, feedback does not necessarily come from humans directly. The outcome of a theorem-proof engine~\cite{jiang2021lisa,wu2021lime,wu2022insights} or the execution result of model-generated scripts can also be used as the reward source~\cite{polu2020generative}.




\subsection{Multi-modal Mathematical Reasoning}

In recent years, there has been growing interest in multi-modal mathematical reasoning, which involves using multiple sources of information, such as text, tables, natural images, and diagrams
~\cite{kahou2017figureqa,kafle2018dvqa,lu2021iconqa,lu2022dynamic}. 
However, currently available datasets in this domain tend to be small~\cite{zhao2022multihiertt}, generated from templates~\cite{kahou2017figureqa}, or focus on specific topics~\cite{lu2021inter,chen2022unigeo}. One line of current research involves applying VQA-based frameworks to analyze figures and plots, but this approach can result in significant semantic gaps due to the fact that most VQA models are trained on natural images. 
One potential direction for future work is to enhance the ability of multi-modal mathematical reasoning systems to tackle more complex and realistic problems. This may involve creating unified models for interpreting and integrating different modalities, as well as developing better evaluation benchmarks to assess the performance of these systems.



\section{Conclusion}
In this paper, we present a comprehensive survey of deep learning for mathematical reasoning. We review the various tasks,  datasets, and deep learning approaches. We also identify several gaps in the existing datasets and methods. Finally, we outline directions for future research and highlight the potential for further exploration in this field. Our goal with this paper is to provide a comprehensive and useful resource for readers interested in the development of deep learning for mathematical reasoning. To aid in this effort, we have created a reading list that will be continually updated in a GitHub repository at \href{https://github.com/lupantech/dl4math}{https://github.com/lupantech/dl4math}.


\section*{Limitations}
One limitation of our survey work is that it is focused on the intersection of mathematical reasoning and deep learning over the past decade, which may not encompass the entire field and its history. Additionally, our evaluation of existing benchmarks and methods is based on a curated set of papers and may not fully represent the state of the art in the field. Furthermore, due to the fast-paced nature of the field, our survey may not reflect the latest developments and advancements which may have come out close to or after the survey was conducted. Despite these limitations, our survey still provides a valuable overview of the current state and key trends in the field of mathematical reasoning and deep learning, and can serve as a valuable resource for researchers and practitioners working in this field.

\section*{Broader Impact}
Our survey paper on the intersection of mathematical reasoning and deep learning has the potential to significantly impact the field of artificial intelligence. By providing a comprehensive overview of the key tasks, datasets, and methods that have been developed in the past decade, we give researchers and practitioners a clear understanding of the current state-of-the-art and help them make informed decisions about their own research. Additionally, by evaluating existing benchmarks and methods and discussing future research directions, we aim to identify gaps in the current state of the art and guide future research and development efforts towards more advanced and effective mathematical reasoning systems. Overall, our survey has the potential to contribute to the advancement of mathematical reasoning and deep learning, and have a profound impact on machine learning and natural language processing.

\bibliography{custom}
\bibliographystyle{acl_natbib}
\nocite{*}

\newpage
\appendix

\section{Mathematical Reasoning  Datasets}
\label{sec:dataset}

\begin{figure}[t!] 
\centering
\includegraphics[width=0.49\textwidth]{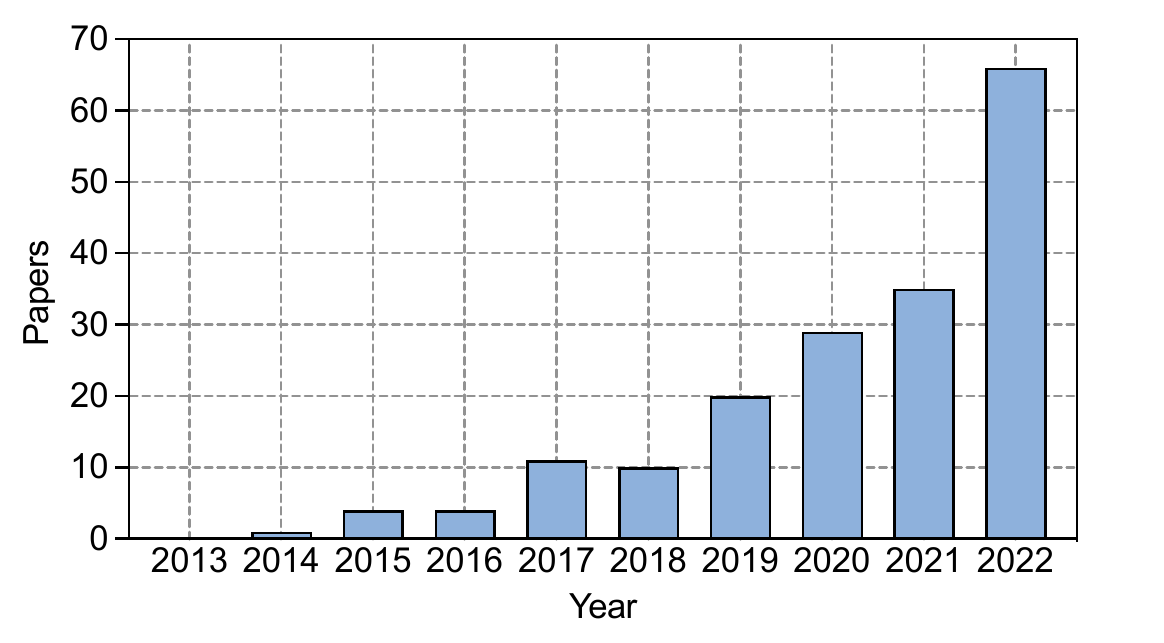}
\vspace{-6mm}
\caption{Estimated counts of annually published papers on deep learning for mathematical reasoning. This field has been experiencing rapid growth since 2018.}
\vspace{-4mm}
\label{fig:papers}
\end{figure}

In this section, we will examine the various datasets currently available for the study of mathematical reasoning using deep learning methods. A summary of the commonly used datasets in this field can be found in \autoref{tab:datasets}.


\subsection{Math Word Problem Solving}
\label{sec:mwp}

Developing algorithms to solve math word problems (MWPs) automatically has been an interest of NLP researchers for decades \cite{feigenbaum1963computers,bobrow1964natural}. A math word problem (also termed an algebraic or arithmetic word problem) describes a brief narrative that involves characters, entities, and quantities. The mathematical relationship of an MWP can be modeled with a set of equations whose solution reveals the final answer to the question. A typical example is shown in \autoref{tab:mwp}. A question involves the four basic arithmetic operations of addition, subtraction, multiplication, and division with single or multiple operation steps. The challenge of MWPs for NLP systems lies in the need for language comprehension, semantic parsing, and multiple mathematical reasoning skills.

Existing MWP datasets cover grade school problems, which are crawled from online learning websites \cite{koncel2015parsing}, collected from textbooks, or manually annotated by human workers \cite{patel2021nlp}. Early math word problem datasets are relatively small or limited to a small number of operation steps \cite{hosseini-etal-2014-learning, kushman2014learning,roy2015reasoning}. Some recently curated datasets aim to increase problem diversity and difficulty levels. For example, Ape210K \cite{zhao2020ape210k} consists of 210k elementary math word problems, which is the largest publicly available. The problems in GSM8K \cite{cobbe2021training} can involve up to 8 steps to solve. SVAMP \cite{patel2021nlp} is a benchmark that tests the robustness of deep learning models to math word problems with simple variations. More recently built datasets involve modalities beyond text.
For example, IconQA \cite{lu2021iconqa} provides an abstract diagram as a visual context, while TabMWP \cite{lu2022dynamic} provides a tabular context for each problem. 

Most MWP datasets provide annotated equations as a rationale for the solution (e.g., \autoref{tab:mwp}).
To improve the performance and interpretability of the learned solvers, MathQA \cite{tafjord2019quarel} is annotated with precise operation programs, and MathQA-Python \cite{austin2021program} is provided with specific Python programs instead. Another line of datasets annotates the problems with multi-step natural language solutions that are regarded as more human-readable \cite{ling2017program,cobbe2021training,lu2022dynamic}. 
Lila \cite{Mishra2022Lila} annotates many of the previously mentioned MWP datasets with Python program rationales.

\subsection{Theorem Proving}
\label{sec:tp}

%

Recently, there has been increased interest in using language models for theorem proving in formal \textit{interactive theorem provers} (ITP) (e.g.,~\citet{polu2020generative,han2021proof,polu2022FormalMS,jiang2022thor,jiang2022dsp,lample2022hypertree}).
Example ITPs include Lean~\cite{moura2015lean}, Isabelle~\cite{paulson1994isabelle}, Coq~\cite{barras1999coq}, and Metamath~\cite{metamath}.
To prove a theorem in an ITP, the theorem is stated in the ITP’s programming language, then simplified by generating ``proof steps'' until it is reduced to known facts. 
The result is a sequence of steps that constitutes a verified proof. 

Data sources for neural theorem proving in ITPs include
interactive learning environments that interface with ITPs, and datasets derived from proofs in ITP libraries.
For example, 
CoqGym \cite{yang2019learning} provides an interactive environment and 71K human-written proofs for the Coq ITP. For Isabelle, PISA~\cite{jiang2021lisa} enables interaction and provides a dataset of 183k proofs mined from the Isabelle standard library and Archive of Formal Proofs.
For Lean, LeanStep~\cite{han2021proof} provides a dataset of proof-steps from Lean's mathematical library along with auxiliary tasks, while Lean-Gym~\cite{polu2022FormalMS} provides an interactive REPL.
The miniF2F \cite{zheng2021minif2f} benchmark aims to provide a shared benchmark across ITPs, consisting of 488 problem
statements sourced from mathematical competitions.

Other resources provide proxy environments or tasks.
For example, INT \cite{wu2021int} provide a synthetic proving environment to measure six different types of generalization. \citeauthor{li2020isarstep} construct IsarStep using the Isabelle Archive of Formal Proofs, and propose a task of filling in a missing intermediate proposition.
Early applications of deep learning for formal theorem proving focus on selecting relevant premises~\cite{alemi2016deepmath}.

\textit{Informal theorem proving} presents an alternative medium for theorem proving, in which  statements and proofs are written in the mixture of natural language and symbols used in ``standard'' mathematics (e.g., in \LaTeX), and are checked for correctness by humans.
Early work focuses on selecting relevant premises~\cite{ferreira-freitas-2020-premise,ferreira-freitas-2020-natural}. 
\citet{welleck2021naturalproofs} develop NaturalProofs, a large-scale dataset of 32k informal mathematical theorems, definitions, and proofs, and  
provide a benchmark for premise selection via retrieval and generation tasks. \citet{welleck2022naturalprover} adapt NaturalProofs for full proof generation, and provide a human evaluation protocol and proxy automatic metrics.

An emerging area of research aims to combine elements of informal and formal theorem proving.
For example, \citet{wu2022autoformalization} explore translating informal statements into formal statements, while \citet{jiang2022dsp} release a new version of the miniF2F benchmark augmented with informal statements and proofs, which we refer to as \textit{miniF2F+informal}.
\citet{jiang2022dsp} explore translating provided (or generated) informal proofs into formal proofs.

\subsection{Geometry Problem Solving}
\label{sec:gps}
Automated geometry problem solving (GPS) is also a long-standing AI task in mathematical reasoning research \cite{gelernter1960empirical,wen1986basic,chou1996automated,ye2008introduction} and has attracted much attention in recent years. Different from a math word problem, a geometry problem consists of a textual description in natural language and a geometric diagram. As shown in \autoref{fig:gps}, the multimodal inputs describe the entities, attributes, and relationships of geometric elements, and the goal is to find the numeric solution to an unknown variable. GPS is a challenging task for deep learning methods due to the complex skills it requires. It involves the ability to parse multimodal information, perform symbolic abstraction, utilize theorem knowledge, and conduct quantitative reasoning. 


Some early datasets are proposed to facilitate research in this domain \cite{seo2015solving,alvin2017synthesis,sachan2017textbooks,sachan2017learning}. However, these datasets are relatively small or not publicly available, which limits the development of deep learning methods. In response to this limitation,  \citeauthor{lu2021inter} create the Geometry3K dataset, which consists of 3,002 multi-choice geometry problems with unified logic form annotations for the multimodal inputs. More recently, larger-scale datasets such as GeoQA \cite{chen2021geoqa}, GeoQA+ \cite{cao2022augmented}, and UniGeo \cite{chen2022unigeo} 
have been introduced and are annotated with programs that can be learned by neural solvers and executed to obtain the final answers.

\subsection{Math Question Answering}
\label{sec:mathqa}
Numerical reasoning is a core ability within human intelligence and plays an important role in many NLP tasks. Aside from theorem proving and grade-level math word problem solving, there is a wide range of question answering (QA) benchmarks that center around mathematical reasoning. In this work, we refer to these tasks as math question answering (MathQA). A large number of datasets have been presented recently. For example, QuaRel \cite{tafjord2019quarel} is a dataset of diverse story questions that involve 19 different types of quantities. McTaco \cite{ZKNR19} studies temporal commonsense problems, while Fermi \cite{kalyan2021much} studies Fermi problems whose answers can only be approximately estimated.

Recent studies have shown that state-of-the-art mathematical reasoning systems might suffer from brittleness in reasoning, in that the models rely on spurious signals and plug-and-chug calculations in the specific dataset to achieve ``satisfactory'' performance \cite{hendrycks2021measuring,mishra2022numglue}. To address this issue, new benchmarks are proposed from various aspects. The Mathematics dataset \cite{saxton2019analysing} consists of many different types of mathematics problems, covering arithmetic, algebra, probability, and calculus. The dataset allows for measuring the algebraic generalization ability of a model. Similarly, MATH \cite{hendrycks2021measuring} consists of challenging competition mathematics to measure the problem-solving ability of models in complex scenarios. 

Some work 
incorporates tabular contexts in the question inputs. For example, FinQA \cite{chen2021finqa}, TAT-QA \cite{zhu2021tat}, and MultiHiertt \cite{zhao2022multihiertt} collect questions that require both table understanding and numeric reasoning to answer. Others, instead, present large-scale unified benchmarks for mathematical reasoning \cite{mishra2022numglue,Mishra2022Lila,chen2023theoremqa}. NumGLUE \cite{mishra2022numglue} is a multi-task benchmark with the goal of evaluating the performance of models on eight different tasks. \citealt{Mishra2022Lila} push this direction further and presents Lila, which consists of 23 mathematical reasoning tasks, spanning a wide range of mathematics topics, linguistic complexity, question formats, and background knowledge requirements.

\subsection{Other Quantitative Problems}
\label{sec:other_problem}
Numbers are an integral part of our daily lives, and we humans reason with numbers in a variety of tasks, such as understanding news, reports, elections, and markets. This has led many in the community to question whether AI systems can effectively perform quantitative reasoning in everyday scenarios. To this end, various benchmarks have been developed to evaluate the capabilities of AI systems in this area.

Diagrams, such as figures, charts, and plots, are essential media that convey 
large amounts of information in a concise way. FigureQA \cite{kahou2017figureqa}, DVQA \cite{kafle2018dvqa}, MNS \cite{zhang2020machine}, PGDP5K \cite{hao2022pgdp5k}, and GeoRE \cite{yu2021geore}, are released to investigate models' abilities to reason about quantitative relationships among entities grounded in diagrams. NumerSense \cite{lin2020birds}, instead, examines whether and to what extent existing pre-trained language models can induce numerical commonsense knowledge.
EQUATE \cite{ravichander2019equate} formalizes aspects of quantitative reasoning in a natural language inference framework. Quantitative reasoning can appear frequently in specific domains like finance, science, and programming. For instance, the ConvFinQA \cite{chen2022convfinqa} targets  numerical reasoning over financial reports in a conversational question answering format. ScienceQA \cite{lu2022learn} involves numerical reasoning in scientific domains, while P3 \cite{schuster2021programming} studies the function inference ability of deep learning models to find a valid input which makes the given program return \texttt{True}.

\begin{table*}[th!]
\centering
\fontsize{8.0pt}{\baselineskip}\selectfont 
\renewcommand\tabcolsep{6.0pt} 
\renewcommand\arraystretch{0.66} 
\begin{tabular}{lcccccc} 
\toprule
\textbf{Dataset} & \textbf{Task} & \textbf{Size} & \textbf{Input} & \textbf{Output} & \textbf{Rationale} & \textbf{Domain} \\ 
\midrule
\textbf{Verb395} \citeyearpar{hosseini-etal-2014-learning} & MWP & 395 & Question & Number & Equation & Math \\
\textbf{Alg514} \citeyearpar{kushman2014learning} & MWP & 514 & Question & Number & Equation & Math \\
\textbf{IL} \citeyearpar{roy2015reasoning} & MWP & - & Question & Number & Equation & Math \\
\textbf{SingleEQ} \citeyearpar{koncel2015parsing} & MWP & 508 & Question & Number & Equation & Math \\
\textbf{DRAW} \citeyearpar{upadhyay2015draw} & MWP & 1,000 & Question & Number & Equation & Math \\
\textbf{Dolphin1878} \citeyearpar{shi2015automatically} & MWP & 1,878 & Question & Number & Equation & Math \\
\textbf{Dolphin18K} \citeyearpar{huang2016well} & MWP & 18,460 & Question & Number & Equation & Math \\
\textbf{MAWPS} \citeyearpar{koncel2016mawps} & MWP & 3,320 & Question & Number & Equation & Math \\
\textbf{AllArith} \citeyearpar{roy2017unit} & MWP & 831 & Question & Number & Equation & Math \\
\textbf{DRAW-1K} \citeyearpar{upadhyay2017annotating} & MWP & 1,000 & Question & Number & Equation & Math \\
\textbf{Math23K} \citeyearpar{wang2017deep} & MWP & 23,162 & Question & Number & Equation & Math \\
\textbf{AQuA} \citeyearpar{ling2017program} & MWP & 100,000 & Question & Option & Natural language & Math \\
\textbf{Aggregate} \citeyearpar{roy2018mapping} & MWP & 1,492 & Question & Number & Equation & Math \\
\textbf{MathQA} \citeyearpar{amini2019mathqa} & MWP & 37,297 & Question & Number & Program & Math \\
\textbf{ASDiv} \citeyearpar{miao2020diverse} & MWP & 2,305 & Question & Number & Equation & Math \\
\textbf{HMWP} \citeyearpar{qin2020semantically} & MWP & 5,470  & Question & Number & Equation & Math \\
\textbf{Ape210K} \citeyearpar{zhao2020ape210k} & MWP & 210,488 & Question & Number & Equation & Math \\
\textbf{SVAMP} \citeyearpar{patel2021nlp} & MWP & 1,000 & Question & Number & Equation & Math \\
\textbf{GSM8K} \citeyearpar{cobbe2021training} & MWP & 8,792 & Question & Number & Natural language & Math \\
\textbf{IconQA} \citeyearpar{lu2021iconqa} & MWP & 107,439 & Figure+Question & Option+Text span & \xmark & Math \\
\textbf{MathQA-Python} \citeyearpar{austin2021program} & MWP & 23,914 & Question & Number & Python program & Math \\
\textbf{ArMATH} \citeyearpar{alghamdi2022armath} & MWP & 6,000 & Question & Number & Equation & Math \\
\textbf{TabMWP} \citeyearpar{lu2022dynamic} & MWP & 38,431 & Table+Question & Option+Number & Natural language & Math \\
\midrule
\textbf{MML} \citeyearpar{grabowski2015four} & TP & 57,882 & Statement & Proof steps & \xmark & Math \\
\textbf{HolStep} \citeyearpar{kaliszyk2017holstep} & TP & 2,209,076 & Statement & Proof steps & \xmark & Math \\
\textbf{Gamepad} \citeyearpar{huang2018gamepad} & TP & - & Statement & Proof steps & \xmark & Math \\
\textbf{CoqGym} \citeyearpar{yang2019learning} & TP & 71,000 & Statement & Proof steps & \xmark & Math \\
\textbf{HOList} \citeyearpar{bansal2019holist} & TP & 29,462 & Statement & Proof steps & \xmark & Math \\
\textbf{IsarStep} \citeyearpar{li2020isarstep} & TP & 860,000 & Statement & Proof steps & \xmark & Math \\
\textbf{PISA} \citeyearpar{jiang2021lisa} & TP & 183,000 & Statement & Proof steps & \xmark & Math \\
\textbf{INT} \citeyearpar{wu2021int} & TP & - & Statement & Proof steps & \xmark & Math \\
\textbf{NaturalProofs} \citeyearpar{welleck2021naturalproofs} & TP & 32,000 & Statement & Proof steps & \xmark & Math \\
\textbf{NaturalProofs-Gen} \citeyearpar{welleck2022naturalprover} & TP & 14,500 & Statement & Proof steps & \xmark & Math \\
\textbf{miniF2F} \citeyearpar{zheng2021minif2f} & TP & 488 & Statement & Proof steps & \xmark & Math \\
\textbf{miniF2F+informal} \citeyearpar{jiang2022dsp} & TP & 488 & Statement & Proof steps & \xmark & Math \\
\textbf{LeanStep} \citeyearpar{han2021proof} & TP & 21,606,000 & Statement & Proof steps & \xmark & Math \\
\midrule
\textbf{GEOS} \citeyearpar{seo2015solving} & GPS & 186 & Figure+Question & Option & \xmark & Geometry \\
\textbf{GeoShader} \citeyearpar{alvin2017synthesis} & GPS & 102 & Figure+Question & Number & \xmark & Geometry \\
\textbf{GEOS++} \citeyearpar{sachan2017textbooks} & GPS & 1,406 & Figure+Question & Number & \xmark & Geometry \\
\textbf{GEOS-OS} \citeyearpar{sachan2017learning} & GPS & 2,235 & Figure+Question & Option & Demonstration & Geometry \\
\textbf{Geometry3K} \citeyearpar{lu2021inter} & GPS & 3,002 & Figure+Question & Option & Logical form & Geometry \\
\textbf{GeoQA} \citeyearpar{chen2021geoqa} & GPS & 4,998 & Figure+Question & Option & Program & Geometry \\
\textbf{GeoQA+} \citeyearpar{cao2022augmented} & GPS & 12,054 & Figure+Question & Option & Program & Geometry \\
\textbf{UniGeo} \citeyearpar{chen2022unigeo} & GPS/TP & 14,541 & Figure+Question & Option & Program & Geometry \\
\midrule
\textbf{Quarel} \citeyearpar{tafjord2019quarel} & MathQA & 2,771 & Question & Option & Logical form & Math \\
\textbf{McTaco} \citeyearpar{ZKNR19} & MathQA & 13,225 & Text+Question & Option & \xmark & Time \\
\textbf{DROP} \citeyearpar{dua2019drop} & MathQA & 96,567 & Passage+Question & Number+Text span & \xmark & Math \\
\textbf{Mathematics} \citeyearpar{saxton2019analysing} & MathQA & 2,010,000 & Question & Free-form & Number & Math \\
\textbf{FinQA} \citeyearpar{chen2021finqa} & MathQA & 8,281 & Text+Table+Q & Number & Program & Finance \\
\textbf{Fermi} \citeyearpar{kalyan2021much} & MathQA & 11,000 & Question & Number & Program+Fact & Math \\
\textbf{MATH} \citeyearpar{hendrycks2021measuring} & MathQA & 12,500 & Question & Number & Natural language & Math \\
\textbf{TAT-QA} \citeyearpar{zhu2021tat} & MathQA & 16,552 & Text+Table+Q & Number+Text span & \xmark & Finance \\
\textbf{AMPS} \citeyearpar{hendrycks2021measuring} & MathQA & 5,000,000 & Question & - & \LaTeX & Math \\
\textbf{MultiHiertt} \citeyearpar{zhao2022multihiertt} & MathQA & 10,440 & Text+Table+Q & Number+Text span & Expression & Finance \\
\textbf{NumGLUE} \citeyearpar{mishra2022numglue} & MathQA & 101,835 & Text+Question & Number+Text span & \xmark & Math \\
\textbf{Lila} \citeyearpar{Mishra2022Lila} & MathQA & 134,000 & Text+Question & Free-form & Python program & Math \\
\midrule
\textbf{FigureQA} \citeyearpar{kahou2017figureqa} & VQA & 1,000,000+ & Figure+Question & Binary & \xmark & Math \\
\textbf{DVQA} \citeyearpar{kafle2018dvqa} & VQA & 3,487,194 & Figure+Question & Text span & Number+Text span & Math \\
\textbf{DREAM} \citeyearpar{sun2019dream} & ConvQA & 10,197 & Dialog+Question & Option & \xmark & Math \\
\textbf{EQUATE} \citeyearpar{ravichander2019equate} & NLI & - & Premise+Hypothesis & Binary & \xmark & Math \\
\textbf{NumerSense} \citeyearpar{lin2020birds} & Filling & 13,600 & Masked question & Word & \xmark & Math \\
\textbf{MNS} \citeyearpar{zhang2020machine} & IQ Test & - & Figure & Number & \xmark & Math \\
\textbf{P3} \citeyearpar{schuster2021programming} & Puzzle & 397 & Text & Program & \xmark & Math \\
\textbf{NOAHQA} \citeyearpar{zhang2021noahqa} & ConvQA & 21,347 & Dialog+Question & Text span & Reasoning graph & Math \\
\textbf{ConvFinQA} \citeyearpar{chen2022convfinqa} & ConvQA & 3,892 & Report+Dialog+Q & Number & Expression & Math \\
\textbf{PGDP5K} \citeyearpar{hao2022pgdp5k} & Parsing & 5,000 & Figure+Question & Number & \xmark & Geometry \\
\textbf{GeoRE} \citeyearpar{chen2022unigeo} & Parsing & 12,901 & Figure+Question & Number & \xmark & Geometry \\
\textbf{ScienceQA} \citeyearpar{lu2022learn} & VQA & 21,208 & Context+Question & Option & Natural language & Science \\

\bottomrule
\end{tabular}
\caption{A summarization of mathematical reasoning datasets.}
\label{tab:datasets}
\end{table*}

\begin{table*}[t!]
\centering
\fontsize{7.8pt}{\baselineskip}\selectfont 
\renewcommand\tabcolsep{1.8pt} 
\renewcommand\arraystretch{0.75} 
\begin{tabular}{lccccccp{0.28\textwidth}} 
\toprule
\textbf{Paper} & \textbf{Task} & \textbf{Problem} & \textbf{Network} & \textbf{Encod} & \textbf{Decod} &\textbf{ATT} & \textbf{Description} \\ 
\midrule
\textbf{DNS}~\cite{wang2017deep} & MWP &  Generation & Seq2Seq & GRU & LSTM & \xmark & The first deep MWP solver \\
\textbf{AnsRat}~\cite{ling2017program} & MWP &  Generation & Seq2Seq & LSTM & LSTM & \xmark & Trained with staged back-propagation \\
\textbf{Math-EN}~\cite{wang2018translating} & MWP &  Generation & Seq2Seq & BiLSTM & LSTM & \cmark & A standard Seq2Seq model with attention \\
\textbf{CASS}~\cite{huang2018neural} & MWP & Generation & Seq2Seq & BiGRU & BiGRU & \cmark & Copy and alignment with RL \\
\textbf{S-Aligned}~\cite{chiang2019semantically} & MWP &  Generation & Seq2Seq & BiLSTM & LSTM & \cmark & Operating symbols \\
\textbf{T-RNN}~\cite{wang2019template} & MWP &  Generation & Seq2Seq & BiLSTM & BiLSTM & \cmark & Predicting a tree-structure math template
\\
\textbf{GROUP-ATT}~\cite{li2019modeling} & MWP &  Generation & Seq2Seq & BiLSTM & LSTM & \cmark & Group attention \\
\textbf{SMART}~\cite{hong2021smart} & MWP &  Generation & Seq2Seq & - & - & \xmark & Explicitly incorporating values \\
\textbf{SelfAtt}~\cite{robaidek2018data} & GPS & Classification & Seq2Seq & BiLSTM & - & \cmark & Multi-hop self-attention \\
\textbf{QuaSP+}~\cite{tafjord2019quarel} & MathQA & Generation & Seq2Seq & BiLSTM & LSTM & \xmark & Adopting attributed grammar \\
\midrule
\textbf{AST-Dec}~\cite{liu2019tree} & MWP &  Generation & Seq2Tree & BiLSTM & Tree & \cmark & Using prefix order decoding
\\
\textbf{GTS}~\cite{xie2019goal} & MWP &  Generation & Seq2Tree & BiGRU & Tree & \cmark & A goal-driven tree-structured approach \\
\textbf{KA-S2T}~\cite{wu2020knowledge} & MWP &  Generation & Seq2Tree & BiLSTM & Tree & \cmark & A knowledge-aware method \\
\textbf{TSN-MD}~\cite{zhang2020teacher} & MWP & Generation & Seq2Tree & BiGRU & Tree & \cmark & A teacher-student network \\
\textbf{T-LSTM}~\cite{zaporojets2021solving} & MWP &  Generation & Seq2Tree & BiLSTM & Tree & \xmark & A child-sum tree-LSTM model \\
\textbf{NT-LSTM}~\cite{zaporojets2021solving} & MWP &  Generation & Seq2Tree & BiLSTM & Tree & \xmark & An N-ary tree-LSTM model  \\
\textbf{NS-Solver}~\cite{qin2021neural} & MWP & Generation  & Seq2Tree & BiGRU & Tree & \cmark & A neural-symbolic solver with programs \\
\textbf{NumS2T}~\cite{wu2021math} & MWP &  Generation & Seq2Tree & BiLSTM & Tree & \cmark & Explicitly incorporating values \\
\textbf{HMS}~\cite{lin2021hms} & MWP &  Generation & Seq2Tree & 
GRU & Tree & \cmark & A word-clause-problem encoder \\
\textbf{LBF}~\cite{hong2021learning} & MWP &  Generation & Seq2Tree & BiGRU & Tree & \cmark & A learning-by-fixing (LBF) framework \\
\textbf{Seq2DAG}~\cite{cao2021bottom} & MWP &  Generation & Seq2Graph & GRU & Graph & \xmark & A direct acyclic graph (DAG) structure \\
\textbf{Graph2Tree}~\cite{zhang2020graph} & MWP &  Generation & Graph2Tree & Graph & Tree & \xmark & Generating  better  solution expressions \\
\textbf{Multi-E/D}~\cite{shen2020solving} & MWP &  Generation & Graph2Tree & Graph & Tree & \cmark & A graph encoder and a tree-bad decoder \\
\textbf{Graph2Tree}~\cite{li2020graph} & MWP &  Generation & Graph2Tree & Graph & Tree & \cmark & A graph-to-tree neural network \\

\textbf{EEH-G2T}~\cite{wu2021edge} & MWP &  Generation & Graph2Tree & Graph & Tree & \xmark & A hierarchical graph-to-tree model \\
\textbf{ASTactic}~\cite{yang2019learning} & TP & Generation & Tree2Seq & TreeLSTM & GRU & \cmark & Generating tactics as programs \\

\midrule
\textbf{MathDQN}~\cite{wang2018mathdqn} & MWP &  Search & DQN & - & - & \xmark & RL with a deep Q-network (DQN) \\
\textbf{DDT}~\cite{meng2019solving} & MWP & Generation & Transformer  & Trm & Trm & \cmark & A Transformer-based model \\
\textbf{DeepMath}~\cite{alemi2016deepmath} & TP & Classification & CNN & CNN & - & \xmark & The first deep large scale theorem prover \\
\textbf{Holophrasm}~\cite{whalen2016holophrasm} & TP & Classification & BiGRU  & BiGRU & - & \xmark & A neural prover for higher-order logic\\
\textbf{CNNTP}~\cite{loos2017deep} & TP & Classification & CNN  & CNN & - & \xmark & A CNN-based theorem prover \\
\textbf{WaveNetTP}~\cite{loos2017deep} & TP & Classification & WaveNet  & WaveNet & - & \xmark & A WaveNet-based theorem prover \\
\textbf{DeepHOL}~\cite{bansal2019holist} & TP & Generation & WaveNet & WaveNet & - & \xmark & A neural theorem prover with RL \\
\textbf{NGS}~\cite{chen2021geoqa} & GPS & Generation & VQA & LSTM* & LSTM & \cmark & The first deep geometry solver \\
\textbf{PGDPNet}~\cite{zhang2022learning} & Parsing & Generation & GNN  & - & - & \xmark & A neural diagram parser with GNN \\
\bottomrule
\end{tabular}
\caption{A summarization of deep neural network models for mathematical reasoning. \textbf{Encod}: encoder, \textbf{Decod}: decoder, \textbf{ATT}: Attention. LSTM*: ResNet + LSTM, Trm: Transformer}
\label{tab:network}
\end{table*}

\end{document}